\documentclass[10pt,twocolumn,letterpaper]{article}

\usepackage[pagenumbers]{cvpr} % To force page numbers, e.g. for an arXiv version

\definecolor{cvprblue}{rgb}{0.21,0.49,0.74}
\usepackage[pagebackref,breaklinks,colorlinks,allcolors=cvprblue]{hyperref}

\def\confName{CVPR}
\def\confYear{2027}

\usepackage{hyperref}
\usepackage{xcolor}
\usepackage{url}
\usepackage{amssymb}
\usepackage{graphicx}
\usepackage{booktabs} 
\usepackage{multirow}
\usepackage[table]{xcolor} 
\usepackage{graphicx}
\usepackage{wrapfig} % 必须加载的宏包
\usepackage{lipsum}
\usepackage{graphicx}

\usepackage{subcaption}
\usepackage[misc]{ifsym}
\usepackage{listings}
\newcommand{\SKILL}{\fcolorbox{gray!15}{gray!15}{\texttt{SKILL.md}}}
\newcommand{\program}{\fcolorbox{gray!15}{gray!15}{\texttt{program.py}}}
\newcommand{\gain}[1]{\rlap{\kern2pt\textcolor{red!80!black}{{\scriptsize (#1)}}}}
\title{VOL-Agent: Programming Your Film via Video Orchestration Language Agentic System}

\def\spaces{~}
\author{Hongji Yang\textsuperscript{1,*}, \spaces Yucheng Zhou\textsuperscript{1,*}, \spaces Lingting Zhu\textsuperscript{3,*}, \spaces Shengju Qian\textsuperscript{2},  \spaces Xinye Li\textsuperscript{4}, \\
Huan Zheng\textsuperscript{1}, Dongyue Lu\textsuperscript{5}, Wei Chow\textsuperscript{5}, Xin Wang\textsuperscript{2}, Cheng-zhong Xu\textsuperscript{1}, Jianbing Shen\textsuperscript{1,~\Letter} \\
\textsuperscript{1}{SKL-IOTSC, Department of AIX, University of Macau} \\ 
\textsuperscript{2}{LIGHTSPEED}~~\textsuperscript{3}{HKU}~~\textsuperscript{4}{CUHK}~~\textsuperscript{5}{NUS}\\
{\tt\small {*}~Equal Contribution, {\Letter}~Corresponding author} \\
{\small \url{ https://github.com/um-lab/vol-agent/}}
} 

\begin{document}
\maketitle
% \vspace{-4mm}
\vspace{-3mm}
\begin{abstract}
Recent video foundation models generate high-quality and limited-length clips, but orchestrating them into minute-level, story-coherent films remains an open problem. Existing video agents highly rely on natural-language planning or predefined workflows, which make their production hard to program, customize, and extend. Therefore, we introduce VOL-Agent, a programmable video agent built on a Video Orchestration Language (VOL). With static compilation and verification, followed by runtime dynamic loading, VOL-Agent decouples structural reliability from execution flexibility. Through the skill-driven executor, VOL-Agent dynamically resolves tools and skills from an open-world repository and supports user-defined skills in natural language. Moreover, VOL-Agent is capable of crafting new director skills from learned knowledge and improves itself through feedback to accumulate filmmaking knowledge. To overcome the disconnect between different segments, we apply a structural memory mechanism to record and recall world and entity states, sustaining entity and scene consistency across different shots. Critic feedback supports local revision, while reusable filmmaking knowledge and evidence-backed error–solution experiences can be retained for subsequent tasks. Experiments on three benchmarks show that VOL-Agent generates minute-scale videos with strong overall quality and limited cost.

\end{abstract}
    
\vspace{-2.5mm}
\section{Introduction}
\vspace{-1mm}
% \begin{wrapfigure}{r}{0.6\linewidth}
% \begin{figure}
%     \centering
%     \vspace{-2mm}
%     \includegraphics[width=\linewidth]{figures/CVPR27_motivation.pdf}
%     \includegraphics[width=\linewidth]{figures/motivation-1.pdf}
%     \vspace{-6mm}
%     \caption{\small The comparison of different video models and agents in long video generation. For a fair comparison, MiniMax-H3 and Seedance2.0-fast use the previous output as the reference, and the same critic and re-generation setup as VOL-Agent.}
%     \label{fig:comparison}
%     \vspace{-4mm}
% % \end{wrapfigure}
% \end{figure}

\begin{figure}[htbp]
    \centering
    \vspace{-2mm}
    \begin{subfigure}{1.0\linewidth}
        \centering
        \includegraphics[width=\linewidth]{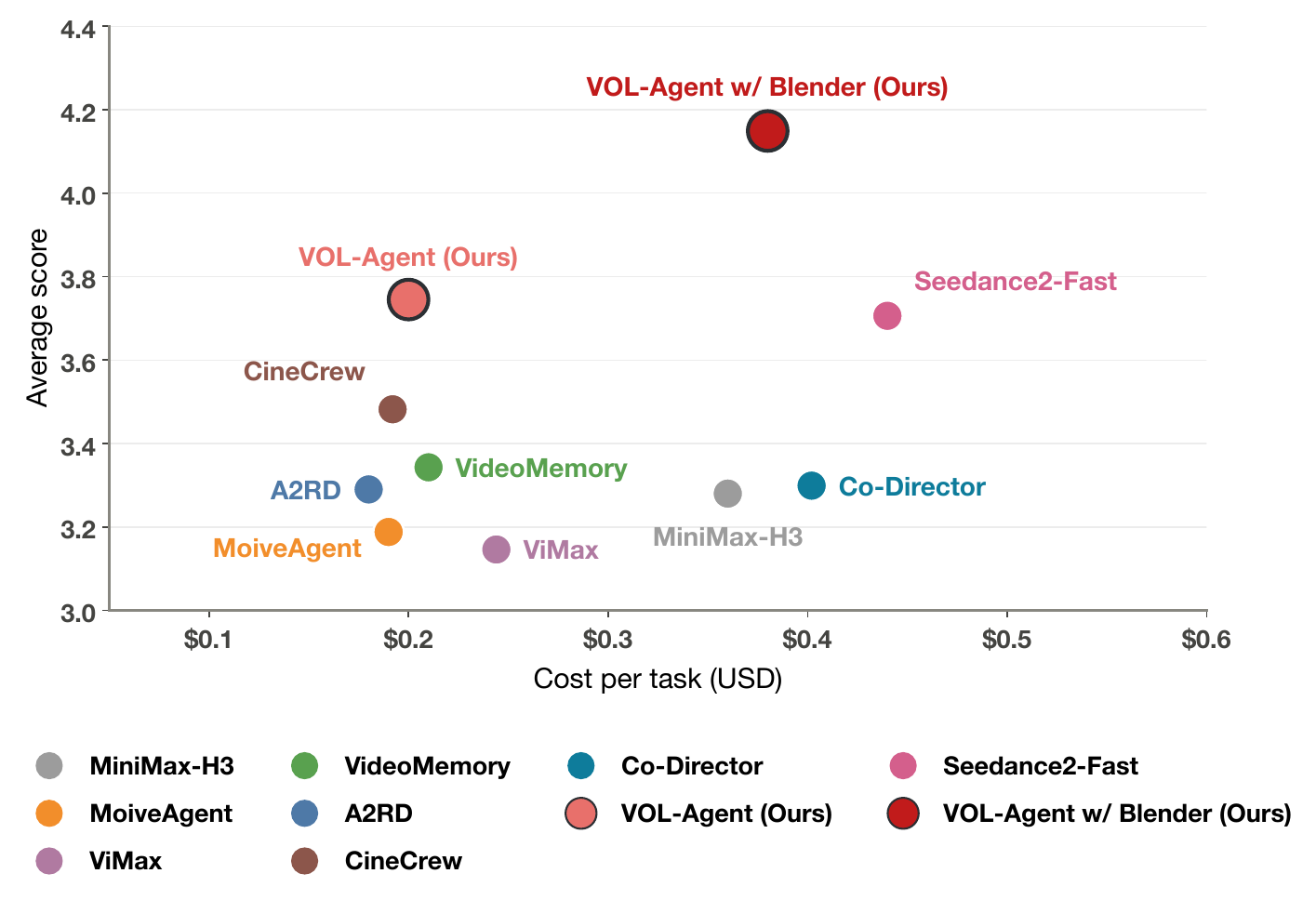}
        % \caption{The cost and performance of different video agents.}
        \vspace{-4.5mm}
        \label{fig:sub_top}
    \end{subfigure}
    \vspace{-10mm}
    \begin{subfigure}{1.0\linewidth}
        \centering
        \includegraphics[width=\linewidth]{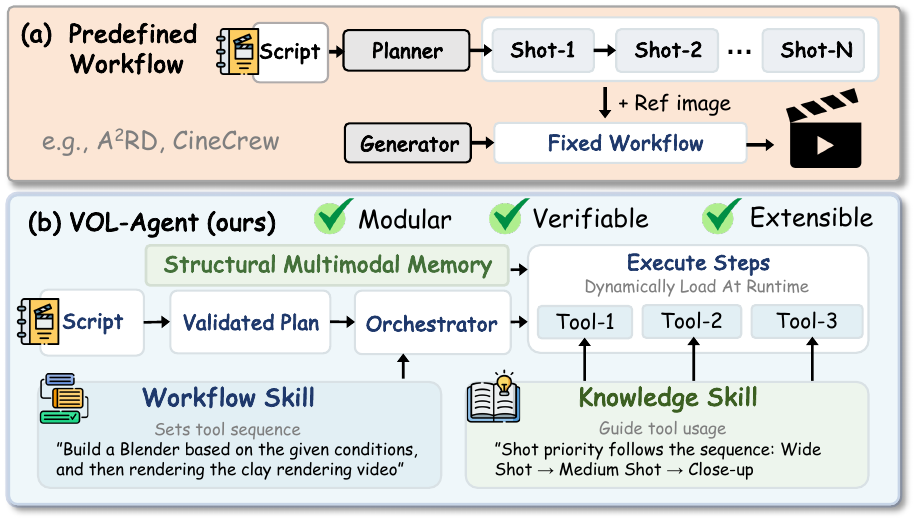}
        % \caption{The comparison of VOL-Agent and the agents using a predefined workflow}
        \label{fig:sub_bottom}
    \end{subfigure}
    \vspace{-2mm}
    \caption{ \small \textbf{Top:} the cost and performance of different video agents; \textbf{Bottom:} the comparison of Predefined workflows versus skill-driven video orchestration in long video generation.}
    \label{fig:motivation}
    \vspace{-6mm}
\end{figure}

Video foundation models have substantially improved the visual quality of generated clips~\citep{yang2025cogvideox,minimax-h3,wan2025wan,seedance2026seedance,veo3, Sora2, team2025kling}.
Early approaches to long video generation adopted open-loop, frame-by-frame or chunk-by-chunk synthesis~\cite{chen2026context, huang2026self, yang2025longlive}.
However, filmmaking requires more than generating a succession of plausible clips. Once generated content is fixed, its errors can propagate into subsequent shots~\cite{zhao2026ultravico, an2026onestory, zhang2025storymem}.
A filmmaking system must therefore maintain the intended story structure and the actual state established by generated shots.

% A script specifies recurring characters, revisited locations, unfinished actions, changes in appearance, and a sequence of camera setups that together convey a story. These requirements must survive many independent generation calls. 
% For example, when a knight kneels before a king and the camera changes angle, the next segment should preserve the knight's identity, kneeling posture, and position relative to the throne, while allowing a different composition. 
% A compelling individual segment does not ensure that these relationships remain coherent across a film.

Agentic video generation~\citep{shi2025animaker,wei2026videoweaver,liu2026cineforge,long20262} addresses this problem by combining narrative planning, visual reference preparation, and iterative synthesis. 
% MovieAgent organizes movie production through hierarchical planning~\citep{wu2025automated}, ViMax coordinates narrative and visual decisions through specialized agents~\citep{huang2026vimax}, and VideoMemory maintains references for recurring visual entities~\citep{zhou2026videomemory}. 
These advances establish the value of planning and memory~\cite{wu2025automated, huang2026vimax, zhou2026videomemory, song2026codirectoragenticgenerativevideo, mu2026scriptneedagenticframework}.
Structured film representations~\citep{chen2026better} further make production elements explicit. Building on these developments, we separate global structural validation from local adaptive execution. Explicit dependencies among entities, scenes, and shots make the structure checkable when plans change or tasks are retried. This motivates a programmable production process: users can inspect and revise dependencies and introduce filmmaking knowledge.

Three challenges motivate our design. First, \textbf{global structure and local execution require different forms of control}. 
The identities of recurring entities, the organization of scenes, and the dependencies between shots should be explicit and checkable before expensive synthesis. In contrast, the concrete procedure for a shot depends on its content, available models, and production preferences. Entangling these decisions in a single free-form interaction makes it difficult to distinguish a structural error from a local rendering failure. Second, \textbf{filmmaking knowledge must be extensible}. Camera conventions, directing styles, prompt formats, and critique criteria vary across productions and change as tools evolve. A fixed workflow cannot anticipate every useful combination, while repeated ad hoc reasoning fails to preserve production experience in a reusable form. Third, \textbf{continuity must follow realized outcomes}. A plan may describe a character entering a room, while the generated clip ends at its doorway. Subsequent generations need the observed state of the video, together with persistent identity and location references, to avoid continuing from an event that never occurred.

We introduce \textbf{VOL-Agent}, a programmable video agent centered on the \textbf{Video Orchestration Language (VOL)}. 
Inspired by the separation between program construction and execution, VOL represents a film as an object graph of entities, scenes, and shot segments. The \textsc{Coding} stage declares this global structure; \textsc{Build} normalizes references, checks dependencies, and produces a static task plan; and \textsc{Run} resolves how each task is executed. Structural validation checks references and dependencies before synthesis, detecting unresolved references and cycles. Runtime evaluation and refinement assess whether generated shots realize the intended content. The compiled plan therefore specifies what the film requires and what each task depends on, without fixing every downstream tool invocation in advance.
As shown in Fig.~\ref{fig:motivation}, unlike hardcoded workflows, VOL-Agent is a modular system in which skills dictate the tool-use order and are injected into each tool, ensuring extensibility with competitive cost and performance.

Execution is driven by an extensible catalog of natural-language skills and registered tools. Workflow skills describe operational sequences, and knowledge skills supply expertise such as shot design, cinematography, and model-specific prompt construction. For each task, the orchestrator selects an applicable workflow, attaches relevant knowledge, and executes the resulting tool sequence. 
The same interface accommodates user-defined skills and director skills learned from reference videos. 
Critic-guided revision further improves videos from feedback, avoiding blind large-scale sampling. 
Furthermore, such successful corrections are stored in an experience library and recalled as a reference for similar cases.
% Critic feedback can revise a shot's prompt and regenerate its candidate; separately invoked maintenance can consolidate transferable lessons into reusable skills and retain supported error--solution pairs as experiences. 
% This form of improvement operates on external knowledge and execution artifacts rather than updating the foundation models.

The memory mechanism~\citep{hu2025memory, zhang2025survey, zhou2026videomemory} of video agent transitions from considering only characters and scenes to incorporating the world state carried over from the previous shot. 
VOL-Agent introduces a structured, multimodal memory mechanism that captures both entities and world state across shots.
We further instantiate a \textbf{clay render workflow} within local execution. The agent constructs executable Blender scenes, renders simplified motion and camera references, reviews them, and conditions final synthesis on the selected clay rendering video and appearance references. Our main contributions are summarized as follows:

\begin{itemize}
    \item We introduce VOL-Agent, separating global compilation and validation (\textsc{Coding} and \textsc{Build}) from adaptive execution (\textsc{Run}) to check dependencies before synthesis while supporting flexible workflows.
    \item We develop a skill-driven agent that dynamically selects tools, supports natural-language skill customization, and crafts and refines filmmaking skills through a learn--generate--evaluate--improve loop to accumulate production knowledge in film.
    \item We propose a structural multimodal memory that combines entity and scene memory with world and entity states observed after each segment, guiding subsequent generation to maintain cross-shot consistency.
     \item We demonstrate that the proposed framework can generate minute-scale videos. Experiments on our benchmark show that our approach achieves strong overall performance, validating the effectiveness of VOL-Agent.
\end{itemize}
\vspace{-2mm}
\section{Related Work}

Video foundation models~\citep{yang2025cogvideox, wan2025wan, minimax-h3} have improved visual quality and motion synthesis. Reference-conditioned models~\citep{Sora2,seedance2026seedance,team2025kling,veo3,sd2.5,guo2024sparsectrl,jiang2025vace} and VLM-guided approaches~\citep{wei2025univideo,chen2026vino,pan2026omniweaving,yang2026cogomnicontrol} further enhance controllability and intent alignment. Producing a complete film from these clip generators requires coordinating multiple calls with script-level planning and cross-shot state management~\cite{chen2026context,huang2026self,yang2025longlive,zhao2026ultravico}. Video agents combine planning, refinement, and memory: MovieAgent~\citep{wu2025automated} uses hierarchical planning, ViMax~\citep{huang2026vimax} coordinates specialized agents, and CineCrew~\citep{chen2026better} introduces FilmDSL for structured production. 
VOL-Agent emphasizes separating global structure and dependency validation from runtime selection of workflows, tools, memories, and skills. 
% Before synthesis, VOL-Agent validates film-wide dependencies among entities, scenes, and shots. 
% At runtime, the orchestrator determines the tool sequence and loads the relevant skills for each shot, using the world and entity states observed in the preceding clip to guide execution.
Test-time scaling~\citep{snell2025scaling} improves outputs through candidate selection and prompt optimization~\citep{gao2025devil,hong2026comic,huang2025planning,long2026vista,zhu2026brandfusion}, while memory preserves information across generations~\citep{li2026storyblender,zuo2026filmworld,shi2025animaker,hu2025memory,zhang2025survey,zhou2026videomemory,li2026forgewmprogressivecausaltraining,wang2026opisinputgroundedbenchmarkmultiobject}. Our framework integrates critic-guided refinement with persistent references and observed shot states to support continuity across a film. 
The full related work discussion is provided in the supplementary material (Sec.~\ref{sec:related_full}).
% The full related work is provided in the Appendix.

\vspace{-2mm}
\section{Method}
  
\subsection{VOL: Video Orchestration Language}
% The core design of the VOL is the global orchestration of planning throughout the whole generation.
% Unlike existing video agents that adopt an autoregressive scheme, dynamically deciding and executing generation step-by-step. In contrast, VOL-Agent strictly disentangles this pipeline. 
Unlike conventional agentic filmmaking pipelines that tightly couple planning with step-by-step generation, VOL-Agent separates global structural planning from local adaptive execution. While existing domain-specific languages, such as FilmDSL in CineCrew, have demonstrated the feasibility of structured film representations, the focus of VOL-Agent is on a compilation-driven orchestration paradigm that explicitly decouples what should be generated from how each generation task is executed.

Given an input script, VOL-Agent accomplishes long-video generation through a three-stage process (\textsc{CODING}, \textsc{BUILD}, and RUN). 
Specifically, VOL-Agent first constructs and statically verifies a global dependency graph of entities, scenes, and shots, ensuring structural validity before expensive video synthesis. Once compiled, each shot is dynamically executed through a skill-driven orchestrator that selects appropriate tools, workflows, and skills at runtime. This separation preserves the validated global structure while allowing local execution strategies to adapt to different generation requirements, available tools, and user-defined skills.
% Inspired by coding agents, VOL-Agent centers around a lightweight domain-specific language (DSL); its primary purpose is to allow the agent to explicitly declare a film's structural framework without simultaneously determining its fine-grained content.
% We argue that a film’s structure is the orchestration of entities, scenes, shots, and their inter-relationships. 
% Given an input script, VOL-Agent accomplishes long-video generation through a three-stage process (\textsc{CODING}, \textsc{BUILD}, and RUN). 
Formally, given a raw script $S_{raw}$, the goal of VOL-Agent is to generate the video sequence $\mathcal{V} = \{V_1, V_2, \dots, V_N\}$, where $N$ denotes the number of shot segments in total. The process can be formulated:
\begin{equation}
P(\mathcal{V} \mid \mathcal{S}_{\mathrm{raw}}) = \underbrace{P(\mathcal{G} \mid \mathcal{S}_{\mathrm{raw}})}_{\mathrm{CODING}} \cdot \underbrace{P(\mathcal{S}_{\mathrm{exec}} \mid \mathcal{G})}_{\mathrm{BUILD}} \cdot \underbrace{P(\mathcal{V} \mid \mathcal{S}_{\mathrm{exec}})}_{\mathrm{RUN}},
\end{equation}

% \begin{equation}
% P(\mathcal{V} \mid \mathcal{S}_{\mathrm{raw}}) = \underbrace{P(\mathcal{G} \mid \mathcal{S}_{\mathrm{raw}})}_{\mathrm{CODING}} \cdot \underbrace{P(\mathcal{S}_{\mathrm{exec}} \mid \mathcal{G})}_{\mathrm{BUILD}} \cdot \underbrace{ \prod^N_{t=1} P(V_t \mid s_t, k_t, M_{t-1})}_{\mathrm{RUN}}
% \end{equation}
where $\mathcal{G}$ is the object graph and $\mathcal{S}_{\mathrm{exec}}=\{s_1, s_2, \dots, s_N\}$ denotes the ordered executable steps after validation and the \textsc{RUN} stage can be formatted as:
\begin{equation}
    P(\mathcal{V} \mid \mathcal{S}_{\mathrm{exec}})= \prod^N_{t=1} P(V_t \mid s_t, k_t, M_{t-1})
\end{equation}
where $M_{t-1}$ denotes the memory of $(t-1)$ shot.

\noindent\textbf{CODING.}
VOL-Agent first parses the raw script and performs global planning (storyboarding). It declares each component as an object graph via a domain-specific programming language. Specifically, the \textsc{CODING} stage formalizes the required characters and prop entities, the target scenes, and shot-level compositions, explicitly mapping the shot content to its associated entity and scene dependencies. 
Crucially, it formulates what content should be generated at each step, leaving explicit definitions of execution tools and workflows for later steps.

% 结构化语言->描述性语言
\noindent\textbf{BUILD.}
The object graph produced by \textsc{CODING} declares the film's skeleton but does not yet contain any of the actual text that will be handed to a generation model.
The stage \textsc{BUILD} turns this skeleton into a shooting script. It first validates whether the graph is well-formatted and every entity, scene, and shot is unique. An executable graph should contain no cycles, and all entities or scenes in the graph must be referenced correctly. This validation is purely mechanical and can be formatted as:
\begin{equation}
    \mathcal{S}_\mathrm{exec}=f_\mathrm{elaborate}(\mathcal{G} | \mathrm{Val}(\mathcal{G})=1),
\end{equation}

Once the graph is valid, VOL-Agent performs \textsc{Elaborate} (driven by LLM) to visit every shot in an order consistent with the dependency graph, and produces the basic storyboard and content while specifying the exact shot transitions, camera changes, and dialogue. The goal of this step is to translate the programming language into a complete, model-agnostic storyboard before dispatching the instructions to the video generation model. 

\noindent\textbf{RUN.}
\textsc{Coding} and \textsc{Build} are responsible for global orchestration and validation. Once the necessary artifacts are ready, the workflow then enters the \textsc{RUN} stage, where each step is locally planned and executed to produce the corresponding visual artifacts. VOL-Agent first invokes the orchestrator to interpret the task of this step and its input/output requirements. The orchestrator then plans the generation workflow, selects the appropriate tools and skills, and organizes them into a to-do list. Then, the tools are executed in the prescribed order until the task is finished.

\begin{figure*}
    \centering
    \vspace{-4mm}
    \includegraphics[width=1.0\linewidth]{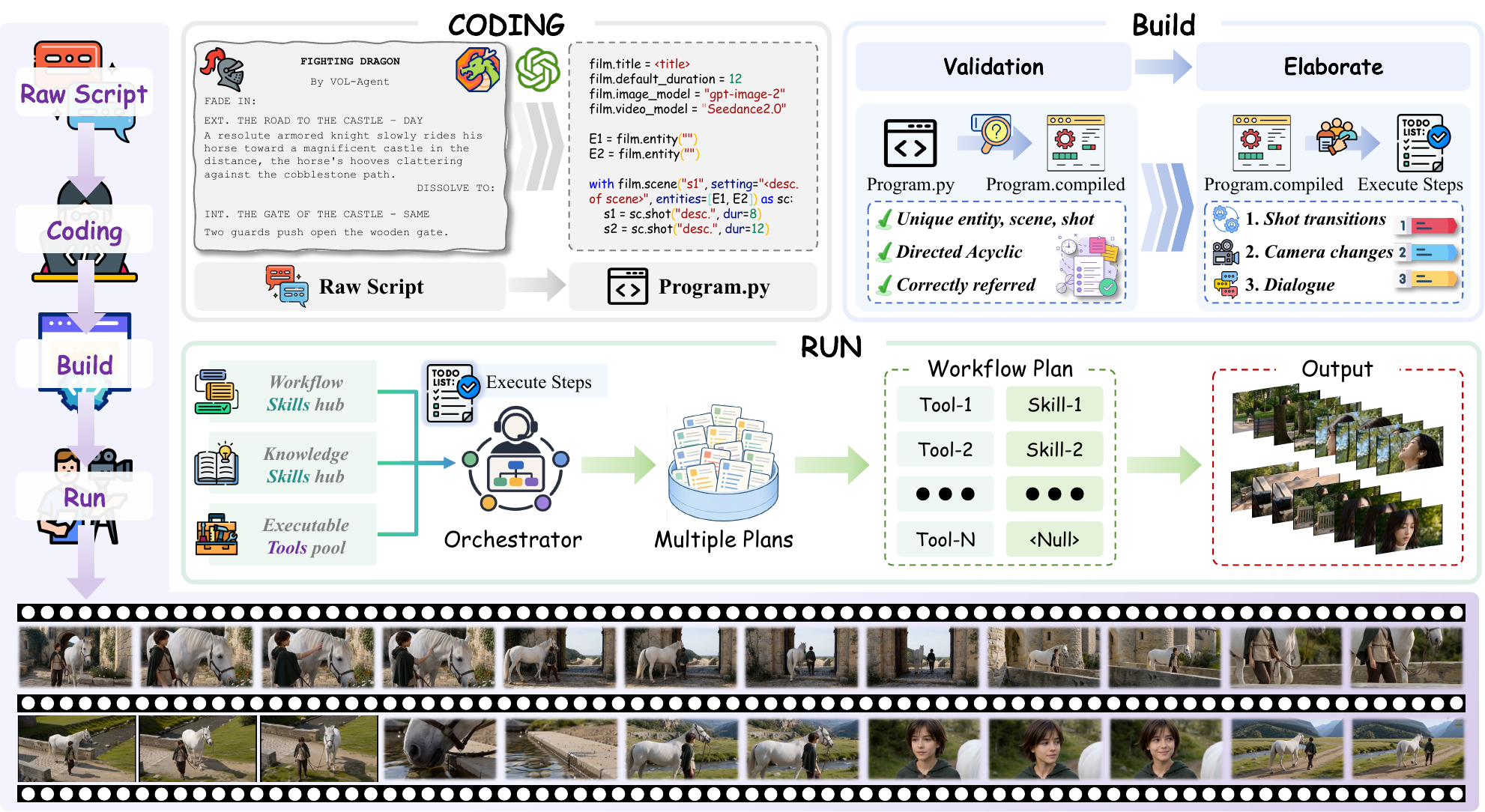}
    \vspace{-6mm}
    \caption{\textbf{The overall framework of the proposed VOL-Agent.} The default workflow of generating long video in VOL-Agent is \textbf{\textsc{CODING}}, \textbf{\textsc{BUILD}}, and \textbf{\textsc{RUN}}. In \textsc{CODING}, the raw script is translated into a program that defines entities, scenes, and shots. In \textsc{BUILD}, the program is validated for unique identifiers, acyclic dependencies, and valid references, then elaborated into execution steps specifying details. In \textsc{RUN}, an orchestrator constructs workflow plans. Each plan specifies the tools and associated skills used to execute the generation steps and produce the final clip.}
    \label{fig:framework}
    \vspace{-4mm}
\end{figure*}

%%%%%%%%%%%%%%%%%%%%%%%%%%%%%%%%%%%%%%%
\subsection{Skill-Driven Execution}
To overcome the rigid adaptability of conventional hard-coded pipelines in long-video generation, we adopt a Skill-Driven Execution Mechanism. 
The core Orchestrator dynamically resolves execution steps and content strategies at runtime by interpreting an open-world repository of human-defined skills, each encapsulated as a natural language specification file (\SKILL ). 

We systematically categorize the skill repository into two functional domains based on how they guide the Orchestrator's decision-making process: \textbf{Workflow Skills} and \textbf{Knowledge Skills}.
Workflow skills define procedural execution topologies and operational sequences. They instruct the Orchestrator on how to order sub-tasks (i.e., what to execute first and what to follow) during a given generation step. 
Knowledge skills inject domain-specific creative priors and professional heuristics into the Orchestrator, which guides the construction of high-quality prompts and generative constraints for downstream video models. These skills incorporate rich cinematographic knowledge, including camera movement conventions, editing rhythms, color aesthetics, and industry-standard production rules.
For $t$-th shot, orchestrator select the skills set from skills hub $\mathcal{K}$:
% $\mathcal{K}=\mathcal{K}_\mathrm{workflow} \cup \mathcal{K}_\mathrm{knowledge}$
% \begin{equation}
%     k_t=\arg \max_{k \in \mathcal{K}} P\big(k \mid s_t \big),
% \end{equation}
% \begin{equation}
%     V_t \sim P(V_t \mid s_t, k_t, M_{t-1}),
% \end{equation}
\begin{equation}
    k_t=f_\text{orch}\big(k \mid s_t \big), {k \in \mathcal{K}}
\end{equation}
where $f_\text{orch}$ denotes the LLM orchestrator in VOL-Agent.

\begin{figure*}
     \centering
     \vspace{-2mm}
    \includegraphics[width=1.0\linewidth]{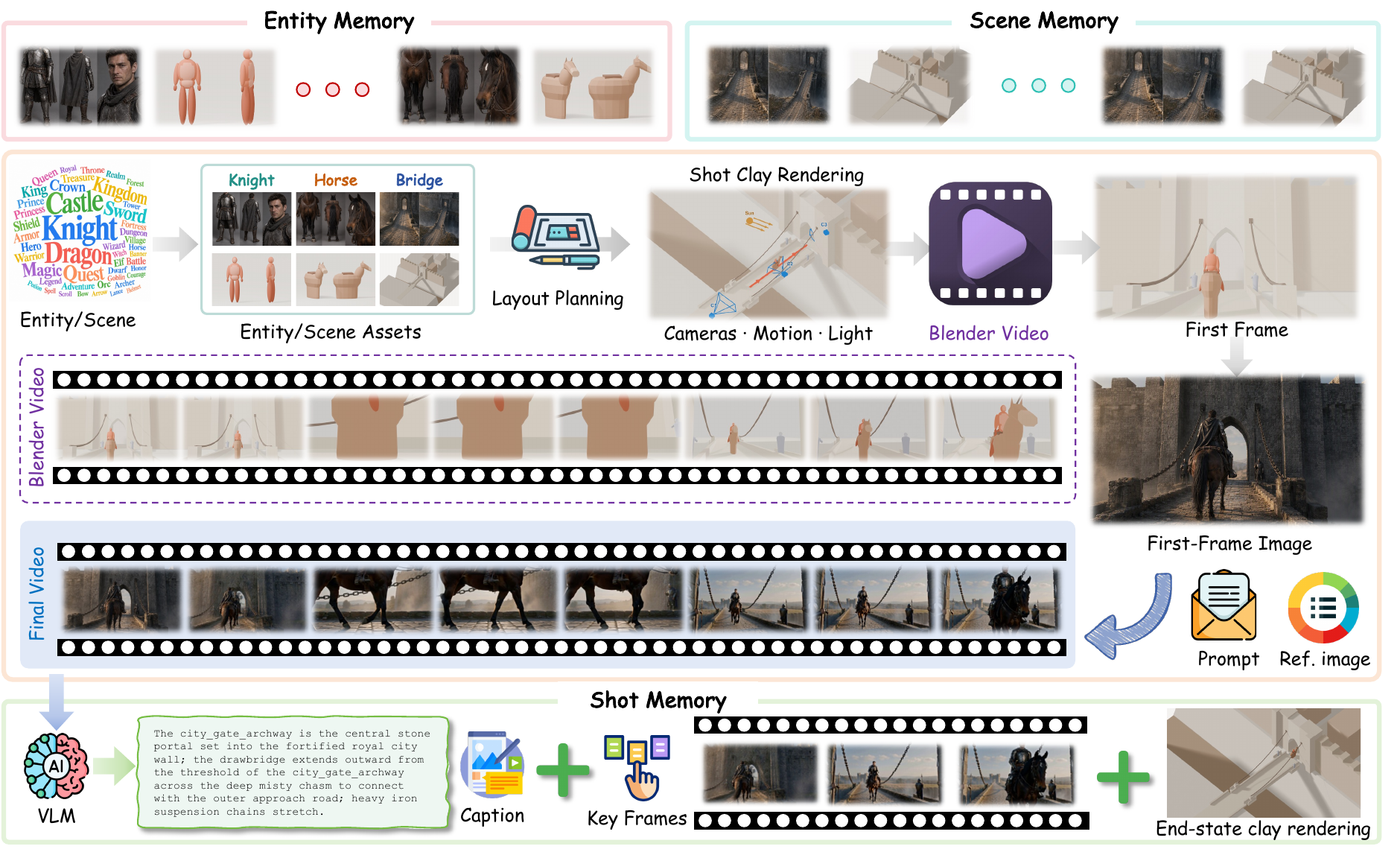}
    \vspace{-6mm}
    \caption{The multi-modal memory mechanism in VOL-Agent. The memory mechanism employs \textbf{Entity Memory}, \textbf{Scene Memory} and \textbf{Shot Memory}. These memories are built from the reference images and optional clay renderings. Reference images provide the visual information of entities or scenes to maintain consistent ID or scene preservation across multiple video generations, and clay rendering ensures structural consistency so that scene layout or entities' positions do not shift due to camera cuts.}
    \vspace{-4mm}
    \label{fig:memory}
\end{figure*}

\subsection{Structural Memory and Conditioning}
\label{sec:memory}
% Reference Memory 提供Entity/Scene的视觉记忆
% Shot Memory （） 提供该shot所发生的时间记忆，方便后续

VOL-Agent maintains a multi-modal memory organized along two axes. The \textit{content} axis contains entity memory $\text{M}^{\text{entity}}$ and scene memory $\text{M}^{\text{scene}}$ that persist throughout the film, whereas shot memory $\text{M}^{\text{shot}}$ is built after each generated segment from the state the video actually reached. The \textit{attribute} axis separates how each entry is represented. For example, a reference image supplies visual reference, and the optional clay rendering~\cite{li2026storyblender, zuo2026filmworld} records 3D proxies and spatial arrangement. VOL-Agent introduces clay rendering through Blender due to its 3D geometry and spatial preservation capability, which is suitable as a memory representation.
The memory can be summarized as follows:
\begin{equation}
    \text{M}_{x} = \bigl( \text{M}_x^{\text{entity}}, \text{M}_x^{\text{scene}}, \text{M}_x^{\text{shot}} \bigr), \quad x \in \{img, clay\},
\end{equation}
The memory can be constructed from reference images and optional clay renderings constructed by Blender.

% % ID-Memory / Scene-Memory / Shot-Memory 
% VOL-Agent maintains three complementary forms of memory during generation: entity-memory $\text{M}^{\text{entity}}$, scene-memory $\text{M}^{\text{scene}}$, and shot-memory $\text{M}^{\text{shot}}_{t}$.
% The three memories serve distinct roles in maintaining identity, environment, and temporal state throughout long-form video generation.
% Entity Memory and Scene Memory serve as persistent reference memories. 
During the RUN stage, VOL-Agent materializes the entities and scenes declared during compilation into visual assets, such as character or subject reference images and scene reference images. Based on the reference images, the optional clay rendering can be built by the LLM-driven Blender. 
These assets are subsequently resolved through the placeholders specified in each shot. In this way, Entity Memory and Scene Memory provide persistent visual references for maintaining the appearance and identity of entities as well as the visual consistency of scenes across shots.

In contrast, shot memory $\text{M}^{\text{shot}}_{t}$ is dynamically constructed from the actual generated videos. 
If clay rendering is available, the LLM places entity clay rendering into the scene clay rendering, with corresponding motion, camera and light, and outputs a video as the conditioning.  
After each shot is generated, a VLM analyzes the resulting video and extracts its current world state $W_{t}$ and entity states $E_{t}$, including the spatial configuration, poses, interactions, and other temporally evolving properties of entities.

The VLM also extracts representative keyframes $K_{t}$ as visual evidence in shot memory. This memory is then used as contextual information when planning subsequent shots, allowing the VOL-Agent to ground future decisions in the actual visual outcomes of previously generated content.
Furthermore, VOL-Agent also saves the clay rendering of the final state and reuses it in the next continuous shot, as it also records the states of the entities and the world.
We define the $t$-th shot memory as:
\begin{equation}
    \text{M}^{\text{shot}}_{t} = \{W_{t}, E_{t}, K_{t}\},
\end{equation}
where $W_t$ denotes the current world state, $E_t$ denotes the current entity state, and $K_t$ contains representative keyframes extracted from the generated video.
During the generation of shot $t$, the orchestrator accesses all three types of memory and produces the final video segment.
\begin{equation}
    \text{V}_{t} \sim P\big(V_t | s_t, M^{\mathrm{entity}}, M^{\mathrm{scene}}, M^{\mathrm{shot}}_{t-1}\big), M^{\mathrm{shot}}_{0}=\varnothing,
\end{equation}
where $S_t$ is the shot specification produced during the \textsc{CODING} and \textsc{BUILD} stage, while $M^{\mathrm{entity}}$ and $M^{\mathrm{scene}}$ provide persistent entity and scene memory that should be referred, respectively. $M_{t-1}^{\mathrm{shot}}$ provides the state of the world established by generated content from the previous shots.

After $V_t$ is generated, a VLM is employed to analyze the video and update the current state, formatted as:
\begin{equation}
    (W_{t}, E_t, K_t) = f_\text{observe}(\text{V}_t),
\end{equation}
Here, $W_t$ captures the global world state, such as spatial configuration, environment status, and persistent object relations, while $E_t$ captures the dynamic state of each entity, including its location, orientation, pose, action, and interactions. $K_t$ provides visual evidence of the realized state through representative keyframes.
These records are enabled when writing the prompt for the next shot to ensure consistency across different shots.

\begin{table*}
    \setlength{\tabcolsep}{8pt}
    \vspace{-3mm}
    \centering
    \resizebox{\textwidth}{!}{
    \setlength{\tabcolsep}{4pt}
    \begin{tabular}{@{}lcccccccc@{}} 
    \toprule
    \multirow{2}{*}{\textbf{Methods} } & \textbf{Entity} & \textbf{Scene} & \textbf{Transition} & \textbf{Semantic} & \textbf{Cinematic} & \textbf{Narrative} & \textbf{Action\&} & \multirow{2}{*}{\textbf{Average}}\\
    ~ & \textbf{Consistency} & \textbf{Consistency} & \textbf{Consistency} & \textbf{Alignment} & \textbf{Aesthetics} & \textbf{Coherence} & \textbf{Staging} \\
    \midrule
    % \rowcolor{gray!10} \multicolumn{10}{Base Model w/ autoregressive} \\
    \rowcolor{gray!10}\multicolumn{9}{c}{\textit{Autoregressive base model}} \\
    \midrule
    MiniMax-H3~\cite{minimax-h3} & 3.539 & 3.573 & 3.051 & 3.206 & 3.468 & 3.210 & 2.917 & 3.280 \\
    Seedance2.0-fast~\cite{seedance2026seedance} & \underline{3.914} & \underline{3.998} & \underline{3.301} & 3.762 & 3.633 & \underline{3.739} & \underline{3.599} & \underline{3.706}  \\
    % Seedance-2.5~\cite{sd2.5} & \textbf{4.332} & \textbf{4.405} & \textbf{3.769} & 4.255 & \textbf{4.326} & 4.206 & \textbf{4.080} & \textbf{4.196}  \\
    \midrule
    \rowcolor{gray!10}\multicolumn{9}{c}{\textit{Agent-based methods}} \\
    \midrule
    MovieAgent~\cite{wu2025automated} & 3.337 & 3.208 & 2.871 & 3.435 & 3.294 & 3.081 & 3.093 & 3.188 \\
    ViMax~\cite{huang2026vimax} & 3.138 & 3.419 & 2.927 & 3.291 & 3.200 & 3.144 & 2.899 & 3.146 \\
    VideoMemory~\cite{zhou2026videomemory} & 3.329 & 3.697 & 3.023 & 3.589 & 3.325 & 3.384 & 3.056 & 3.343 \\
    $\text{A}^{2}\text{RD}$~\cite{long20262} & 3.251 & 3.267 & 3.154 & 3.371 & 3.452 & 3.378 & 3.155 & 3.290 \\
    ScriptAgent~\cite{mu2026scriptneedagenticframework} & 3.060 & 3.179 & 2.239 & 2.524 & 2.825 & 2.171 & 2.448 & 2.635 \\
    CineCrew~\cite{chen2026better} & 3.769 & 3.145 & 3.100 & \underline{3.777} & \underline{3.678} & 3.525 & 3.378 & 3.482 \\
    Co-Director~\cite{song2026codirectoragenticgenerativevideo} & 3.688 & 3.550 & 2.708 & 3.305 & 3.508 & 3.185 & 3.146 & 3.299 \\
    \midrule
    \textbf{VOL-Agent (Ours)} & 3.928 & 3.995 & 3.263 & 3.886 & 3.789 & 3.834 & 3.521 & 3.745 \\
    \textbf{VOL-Agent w/ Blender (Ours)} & \textbf{4.299} & \textbf{4.219} & \textbf{3.756} & \textbf{4.270} & \textbf{4.220} & \textbf{4.248} & \textbf{4.029} & \textbf{4.149} \\
    \bottomrule
    \end{tabular}}
    \vspace{-3mm}
    \caption{The comparison of VOL-Agent with other video agent methods on our benchmark (higher is better).}
    \label{tab:main_exp}
    \vspace{-2mm}
\end{table*}

\subsection{Self-Crafting and Self-improvement via Skill}
\noindent\textbf{Self-Crafting} 
VOL-Agent supports dynamic SKILL-Crafting, which enables the system to learn and construct specialized director skills from websites or provided film videos. These crafted skills are dynamically mounted onto the knowledge base in real time, allowing the Agent to invoke them at specific execution steps in generation.

Formally, the meta-process of skill crafting is itself encapsulated as a meta-skill (defined via a structured \SKILL).
It empowers the proposed VOL-Agent to autonomously acquire and synthesize new director skills, while providing a coupled critic (director dimension) in the same file. 
This self-defined critic establishes fine-grained criteria that guide the Agent on multi-aspect evaluation during the iterative refinement process.
\begin{equation}
    k_\text{director} = f_\text{craft}(\text{V}_\text{ref}, k_\text{craft}),
\end{equation}
where $k_\text{craft}$ denotes the meta crafting skill and $\text{V}_\text{ref}$ and $k_\text{director}$ denote the reference and the output director skill.

% refinement 修改video prompt，refine该shot，以及提升skills
\noindent\textbf{Self-improvement} To enable continuous self-improvement, VOL-Agent introduces a closed-loop process of learning, generation, evaluation, and improvement. A Critic Agent evaluates generated videos according to user-defined or task-specific evaluation skills, identifying production errors and providing concrete revision targets. Based on this feedback, the agent refines the active skills and consolidates the newly discovered knowledge.
The resulting knowledge is stored at two complementary levels. Generalizable improvements are incorporated into skills, progressively improving the agent's reusable production capabilities, while concrete error–solution pairs are retained as experiences in an experience repository. When similar errors arise in future productions, the agent retrieves relevant experiences to directly reuse previously validated solutions, reducing redundant exploration. In this way, VOL-Agent gradually transforms production feedback into an evolving repository of filmmaking skills and experiences of improvement for video generation.

% Critic Agent -> feedback -> adjust skills
To better align the intended generation objective, a critic agent is introduced to evaluate the produced clips and produce feedback.
A critic agent evaluates $V_t$ against task-specific evaluation skills $k_{\mathrm{eval}}\in\mathcal{K}$ across $D$ distinct aesthetic and continuity dimensions and further diagnoses the root causes and provides targeted suggestions.
\begin{equation}
    r_t, \mathcal{F}_t = f_{\mathrm{critic}}\big(V_t, s_t, M_{t-1} \mid k_\mathrm{eval}\big),
\end{equation}
where $r_t = [r^1_t,r^2_t,\dots r^D_t] \in \mathrm{R}^D$ denotes the multi-dimensional scores, and $\mathcal{F}_t$ represents the diagnosis and feedback from critic agent. 
The $f_{\mathrm{critic}}$ denotes the critic VLM in the workflow.
The $k_{\mathrm{eval}}$ contains evaluation dimensions with intent fulfillment, narrative rhythm, cinematic aesthetics, scene match, and temporal coherence. 

The original prompt and inputs, together with the critic's feedback, are then fed into a prompt-rewrite model that revises the prompt. The refined prompt is subsequently used to re-execute the final video generation.
\begin{equation}
    s^{(m+1)}_t = f_{\mathrm{refinement}}\big( s^{(m)}_t, \mathcal{F}_t, M_{t-1}, \xi_{\mathrm{exp}} \big),
    \label{eq:refinement}
\end{equation}
where $\xi_{\mathrm{exp}}$ denotes the experiences of trajectory history of successful modifications in video generation. When encountering analogous generation errors or critique feedback, the Agent retrieves past correction experiences from $\xi_{\mathrm{exp}}$ to guide the current refinement process.

Upon obtaining a validated correction pair $\big( \mathcal{F}_t, \Delta s_t$ where $\Delta s_t = s^{(m)}_t - s^{(0)}_t$, VOL-Agent updates its self-improving memory at two complementary levels:
% \begin{equation}
\begin{align}
    \mathcal{K} \leftarrow \mathcal{K} \cup f_{\mathrm{summarize}} \big ( \{ \mathcal{F}_t, \Delta s_t\} \big), \\
    \xi_{\mathrm{exp}} \leftarrow  \xi_{\mathrm{exp}} \cup \{ (s_t, V_t,  \mathcal{F}_t, s^{(m)}_t) \},
\end{align}
% \end{equation}
% Based on the skill learning, VOL-Agent supports the 
where the $\mathcal{K}$ denotes the skills hub and $f_{\mathrm{summarize}}$ denotes the VLM summarizes the correction.

\begin{table*}
    \vspace{-1mm}
    \setlength{\tabcolsep}{10pt}
    \centering
    \resizebox{\textwidth}{!}{
    \setlength{\tabcolsep}{4pt}
    \begin{tabular}{@{}llcccccccc@{}} 
    \toprule
    \multicolumn{1}{l}{\multirow{2}{*}{\textbf{Benchmark}}} & \multirow{2}{*}{\textbf{Methods} } & \textbf{Entity} & \textbf{Scene} & \textbf{Transition} & \textbf{Semantic} & \textbf{Cinematic} & \textbf{Narrative} & \textbf{Action\&} & \multirow{2}{*}{\textbf{Average}}\\
    \multicolumn{1}{l}{~} & ~ & \textbf{Consistency} & \textbf{Consistency} & \textbf{Consistency} & \textbf{Alignment} & \textbf{Aesthetics} & \textbf{Coherence} & \textbf{Staging} \\
    \midrule
    \multirow{6}{*}{\textbf{ViMaxBench}} & MiniMax-H3~\cite{minimax-h3} & 3.969 & 3.947 & 3.120 & 3.721 & 3.620 & 3.627 & 3.522 & 3.647 \\
    ~ & MovieAgent~\cite{wu2025automated} & 3.557 & 3.725 & 2.794 & 3.587 & 3.511 & 3.408 & 3.263 & 3.406 \\
    ~ & ViMax~\cite{huang2026vimax} & 4.011 & 3.993 & 3.056 & 3.791 & 3.570 & 3.572 & 3.496 & 3.641 \\
    ~ & VideoMemory~\cite{zhou2026videomemory} & 3.858 & 3.893 & 3.155 & 3.686 & 3.676 & 3.510 & 3.422 & 3.600 \\
    ~ & CineCrew~\cite{chen2026better} & 4.302 & 4.279 & 3.070 & 3.909 & 3.788 & 3.778 & 3.718 & 3.835 \\
    ~ & \textbf{VOL-Agent w/ Blender} & \textbf{4.313} & \textbf{4.365} & \textbf{3.736} & \textbf{4.210} & \textbf{4.140} & \textbf{4.271} & \textbf{4.138} & \textbf{4.168} \\
    \midrule
    \multirow{6}{*}{\textbf{MovieBench}} & MiniMax-H3~\cite{minimax-h3} & 3.723 & 3.881 & 3.207 & 3.910 & 3.535 & 3.697 & 3.458 & 3.630 \\
    ~ & MovieAgent~\cite{wu2025automated} & 3.111 & 3.650 & 2.625 & 3.675 & 3.484 & 3.344 & 3.229 & 3.303 \\
    ~ & ViMax~\cite{huang2026vimax} & 3.951 & 3.974 & 3.195 & 3.855 & 3.638 & 3.632 & 3.407 & 3.664 \\
    ~ & VideoMemory~\cite{zhou2026videomemory} & 3.774 & 3.878 & 3.022 & 3.784 & 3.588 & 3.444 & 3.381 & 3.553\\
    ~ & CineCrew~\cite{chen2026better} & 4.236 & \textbf{4.290} & 3.167 & 3.842 & 3.897 & 3.726 & 3.592 & 3.822 \\
    ~ & \textbf{VOL-Agent w/ Blender} & \textbf{4.301} & 4.166 & \textbf{3.629} & \textbf{4.364} & \textbf{4.086} & \textbf{4.267} & \textbf{3.970} & \textbf{4.112} \\
    \bottomrule
    \end{tabular}}
    \vspace{-3mm}
    \caption{The comparison of VOL-Agent with other video agent methods on two public benchmarks. Note that we follow CineCrew to sample 20 narratives from MovieBench.}
    \label{tab:main_exp_on_other_dataset}
    \vspace{-1mm}
\end{table*}

\begin{table*}
    \vspace{-2mm}
    \setlength{\tabcolsep}{10pt}
    \centering
    \resizebox{\textwidth}{!}{
    \setlength{\tabcolsep}{4pt}
    \begin{tabular}{@{}lcccccccc@{}} 
    \toprule
    \multirow{2}{*}{\textbf{Methods} } & \textbf{Entity} & \textbf{Scene} & \textbf{Transition} & \textbf{Semantic} & \textbf{Cinematic} & \textbf{Narrative} & \textbf{Action\&} & \multirow{2}{*}{\textbf{Average}}\\
    ~ & \textbf{Consistency} & \textbf{Consistency} & \textbf{Consistency} & \textbf{Alignment} & \textbf{Aesthetics} & \textbf{Coherence} & \textbf{Staging} \\
    \midrule
    Entity\&Scene memory & 4.208 & 4.108 & 3.252 & 4.078 & 4.128 & 4.036 & 3.988 & 3.971 \\
    + Shot memory & 4.281 & 4.207 & 3.671 & 4.181 & 4.207 & 4.182 & 4.010 & 4.105 \\
    w/o Critic & 3.915 & 3.972 & 3.554 & 4.096 & 4.082 & 4.089 & 3.992 & 3.957 \\
    w/o Experience & 3.998 & 4.074 & 3.582 & 4.103 & 4.063 & 4.109 & 4.027 & 3.993 \\
    w/o VOL & 3.823 & 3.909 & 3.517 & 4.087 & 4.079 & 4.101 & 3.807 & 3.903 \\
    \midrule
    \textbf{VOL-Agent} & \textbf{4.299} & \textbf{4.219} & \textbf{3.756} & \textbf{4.270} & \textbf{4.220} & \textbf{4.248} & \textbf{4.029} & \textbf{4.149} \\
    \bottomrule
    \end{tabular}}
    \vspace{-3mm}
    \caption{The ablation studies of VOL-Agent on the proposed benchmark. }
    \label{tab:ablation}
    \vspace{-3mm}
\end{table*}

% \begin{table}
%     \centering
%     \resizebox{\linewidth}{!}{
%     \setlength{\tabcolsep}{3pt}
%     \begin{tabular}{@{}lcccccccc@{}} 
%     \toprule
%     \textbf{Methods} & \textbf{EC} & \textbf{SC} & \textbf{TC} & \textbf{SA} & \textbf{CA} & \textbf{NC} & \textbf{AS} & \textbf{Avg.} \\
%     \midrule
%     w/o $\text{M}^\text{shot}$. & 4.208 & 4.108 & 3.252 & 4.078 & 4.128 & 4.036 & 3.988 & 3.971 \\
%     % w/ Shot & 4.281 & 4.207 & 3.671 & 4.181 & 4.207 & 4.182 & 4.010 & 4.105 \\
%     w/o Critic & 3.915 & 3.972 & 3.554 & 4.096 & 4.082 & 4.089 & 3.992 & 3.957 \\
%     w/o Exp. & 3.998 & 4.074 & 3.582 & 4.103 & 4.063 & 4.109 & 4.027 & 3.993 \\
%     w/o VOL & 3.823 & 3.909 & 3.517 & 4.087 & 4.079 & 4.101 & 3.807 & 3.903 \\
%     \midrule
%     \textbf{VOL-Agent} & \textbf{4.299} & \textbf{4.219} & \textbf{3.756} & \textbf{4.270} & \textbf{4.220} & \textbf{4.248} & \textbf{4.029} & \textbf{4.149} \\
%     \bottomrule
%     \end{tabular}}
%     \vspace{-3mm}
%     \caption{The ablation studies of VOL-Agent on the proposed benchmark. }
%     \label{tab:ablation}
%     \vspace{-3mm}
% \end{table}

\vspace{-2mm}
\section{Experiment}
\vspace{-1mm}
\subsection{Experimental Settings}
\vspace{-1mm}
We evaluate on three benchmarks: our 20-script benchmark, ViMax-Bench~\cite{huang2026vimax}, and 20 narratives sampled from MovieBench~\cite{wu2025automated}. Agent comparisons share planning, vision, and generation models, with MiniMax-H3~\cite{minimax-h3} producing 720P videos. Each shot uses two video-generation calls: Best-of-2 selection for baselines and one critic-guided revision for VOL-Agent. 
% Full model configurations and benchmark details are provided in the supplementary material (Sec.~\ref{sec:full_settings}).
Full model configurations~\cite{gpt-6.1-sol,gemini-3.7-flash,gpt-image-2} and benchmark details are provided in the Appendix.

An independent Gemini-3.8-Flash evaluator~\cite{gemini-3.8-flash} assigns integer scores from 0 to 5 for seven dimensions: Entity Consistency, Scene Consistency, Transition Consistency, Semantic Alignment, Cinematic Aesthetics, Narrative Coherence, and Action \& Staging. 
% Detailed metric descriptions are provided in the supplementary material (Sec.~\ref{sec:full_metrics}).
Detailed metric descriptions are provided in the supplementary material.

\begin{figure*}
    \centering
    \vspace{-3mm}
    \includegraphics[width=\linewidth]{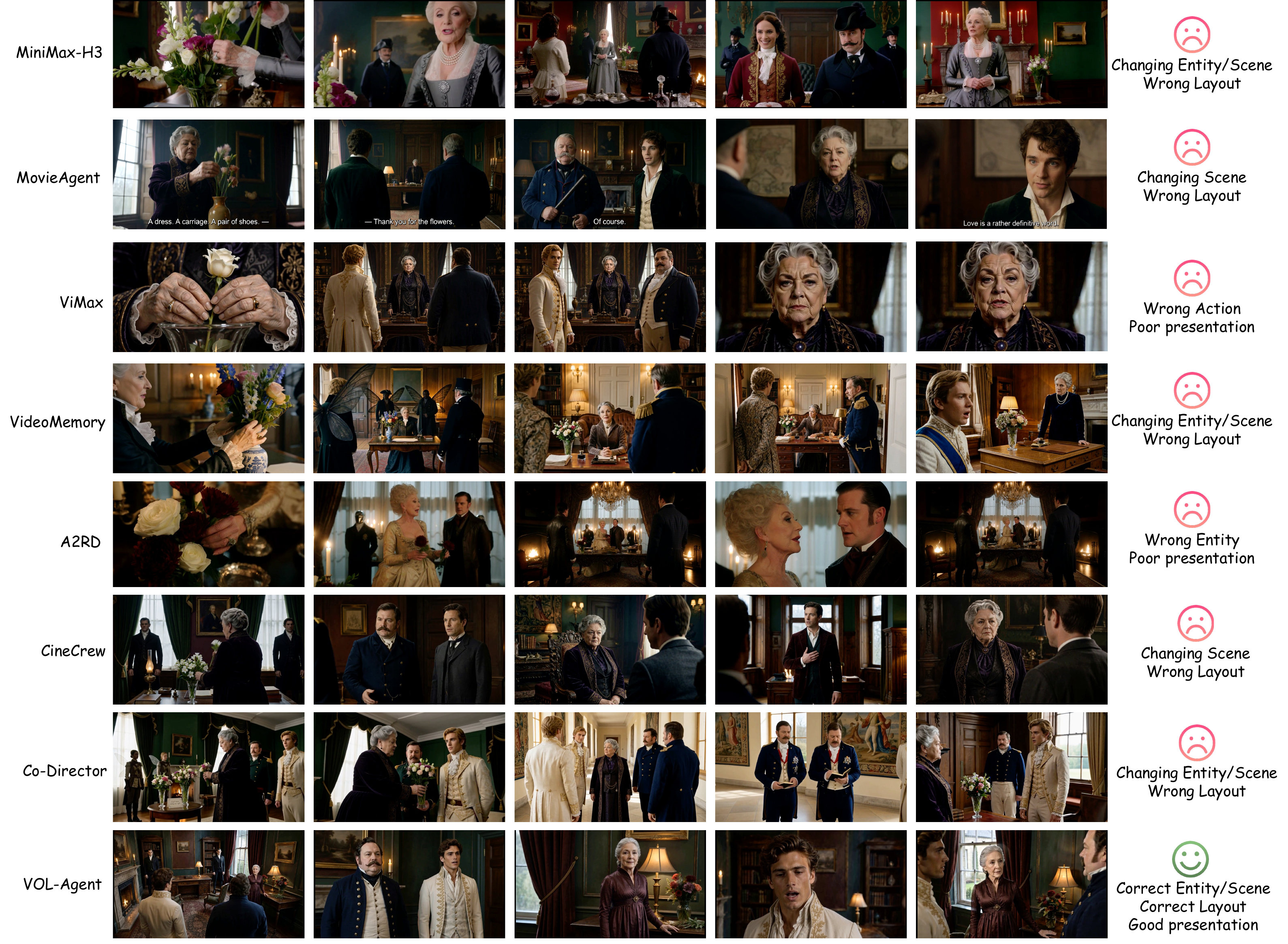}
    \vspace{-7mm}
    \caption{\small Qualitative comparison across different methods.  }
    \label{fig:visual_result2}
    \vspace{-2mm}
\end{figure*}

\vspace{-2mm}
\subsection{Results}
\vspace{-1mm}
\noindent\textbf{Quantitative Results.}
Tab.~\ref{tab:main_exp} shows that VOL-Agent improves overall video quality while maintaining strong entity and scene consistency. Without the Blender workflow, VOL-Agent achieves an average score of 3.745, outperforming CineCrew by 0.263.
These gains are accompanied by higher semantic alignment, cinematic aesthetics, narrative coherence, and action and staging scores. 
Incorporating the Blender-based clay rendering workflow further increases the average score to 4.149, the highest among the compared methods. The improvements are consistent with the role of clay rendering in making spatial arrangements, motion, and camera changes explicit before final synthesis.
We also perform experiments on two public benchmarks. As shown in Tab.~\ref{tab:main_exp_on_other_dataset}, VOL-Agent outperforms CineCrew by 0.333 on ViMaxBench and 0.29 on MovieBench. 
This demonstrates that the proposed VOL-Agent is robust on different public benchmarks.

% 20 3-min
% 20 5-min
% % 10 10-min
% \begin{table*}
%     \setlength{\tabcolsep}{12pt}
%     \centering
%     \resizebox{\textwidth}{!}{
%     \setlength{\tabcolsep}{9pt}
%     \begin{tabular}{@{}lcccccccc@{}} 
%     \toprule
%     \multirow{2}{*}{\textbf{Methods} } & \textbf{Entity} & \textbf{Scene} & \textbf{Transition} & \textbf{Semantic} & \textbf{Cinematic} & \textbf{Narrative} & \textbf{Action\&} & \multirow{2}{*}{\textbf{Average}}\\
%     ~ & \textbf{Consistency} & \textbf{Consistency} & \textbf{Consistency} & \textbf{Alignment} & \textbf{Aesthetics} & \textbf{Coherence} & \textbf{Staging} \\
%     \midrule
%     % \rowcolor{gray!10} \multicolumn{10}{c}{\textit{Base Model}} \\
%     MiniMax-H3 & 3.32 & 3.02 & 4.12 & 4.77 & 4.92 & 4.92 & 4.86 & 4.28 \\
%     Seedance-2.0 \\
%     Seedance-2.5 \\
%     \midrule
%     % \rowcolor{gray!10} \multicolumn{10}{c}{\textit{Agentic System}} \\
%     MovieAgent \\
%     ViMax\\
%     VideoMemory\\
%     $\text{A}^{2}\text{RD}$(MiniMax-H3) & 4.02 & 4.59 & 4.33 & 4.22 & 4.65 & 4.57 & 4.49 & 4.41 \\
%     $\text{A}^{2}\text{RD}$(Seedance-2.0) \\
%     \midrule
%     \textbf{VOL-Agent (MiniMax-H3)} & 4.76 & 4.90 & 4.30 & 4.81 & 4.90 & 4.93 & 4.87 & 4.78 \\
%     \textbf{VOL-Agent (Seedance-2.0)} \\
%     \bottomrule
%     \end{tabular}}
%     \label{tab:main_exp}
    
% \end{table*}

\begin{table}
    \centering
    \resizebox{\linewidth}{!}{
    \setlength{\tabcolsep}{3pt}
    \begin{tabular}{@{}lcccccccc@{}} 
    \toprule
    \textbf{Methods} & \textbf{EC} & \textbf{SC} & \textbf{TC} & \textbf{SA} & \textbf{CA} & \textbf{NC} & \textbf{AS} & \textbf{Avg.} \\
    \midrule
    Baseline & 4.214 & 4.143 & 3.462 & 4.214 & 4.000 & 4.071 & 3.786 & 3.984 \\
    \midrule
    \textit{Best-of-$N$ }& \\
    \textbf{N=2} & 4.267 & 4.178 & 3.662 & 4.273 & 3.857 & 3.929 & 3.792 & 3.994 \\
    N=4 & 4.071 & 4.214 & 3.615 & 4.214 & 3.929 & 4.001 & 3.929 & 3.996 \\
    N=6 & 4.211 & 4.141 & 3.690 & 4.355 & 3.998 & 4.069 & 3.570 & 4.005 \\
    \midrule
    \textit{Refine-$N$} & \\
    \textbf{N=1} & 4.286 & 4.314 & 3.715 & 4.286 & 4.171 & 4.271 & 4.100 & 4.163 \\
    N=2 & 4.312 & 4.355 & 3.767 & 4.307 & 4.215 & 4.284 & 4.141 & 4.197 \\
    N=3 & 4.282 & 4.346 & 3.757 & 4.296 & 4.219 & 4.272 & 4.270 & 4.206 \\
    \bottomrule
    \end{tabular}}
    \vspace{-3.5mm}
    \caption{The results with Best-of-$N$ selection and critic-guided refinement $N$ times on the subset of our benchmark.}
    \label{tab:BoN}
    \vspace{-4mm}
\end{table}

\begin{table}
    \centering
    \resizebox{\linewidth}{!}{
    \setlength{\tabcolsep}{9pt}
    \begin{tabular}{@{}lccc@{}} 
    \toprule
    \textbf{Comparison} & \textbf{Win}($\uparrow$) & \textbf{Tie} & \textbf{Lose}($\downarrow$) \\
    \midrule
    Round-2 vs. No Reference & 97.9\% & ~2.1\% & ~0.0\% \\
    Round-2 vs. Ref. Video as Input & 80.2\% & 14.6\% & ~5.2\% \\
    Round-2 vs. Round-1 & 24.0\% & 57.3\% & 18.7\% \\
    \bottomrule
    \end{tabular}}
    \vspace{-3mm}
    \caption{\small Reference-style alignment of VOL-Agent using different round self-crafting director skills, as ranked by Gemini-3.8-Flash.}
    \label{tab:director}
    \vspace{-5mm}
\end{table}

\noindent\textbf{Qualitative Results.}
Fig.~\ref{fig:visual_result2} compares entity, scene, action, and layout consistency across camera cuts. Competing outputs exhibit identity or scene changes, incorrect staging, and inconsistent layouts, while VOL-Agent better preserves the intended configurations. 
% Additional comparisons and director-style examples are provided in the supplementary material (Sec.~\ref{sec:additional_visual}, Fig.~\ref{fig:visual_result}).
Additional comparisons and director-style examples are provided in the Appendix.

\subsection{Ablation Studies}
We examine the contributions of multimodal memory, critic feedback, and experience in VOL-Agent.
Persistent entity and scene references already yield high entity and scene consistency scores of 4.208 and 4.108. Adding shot memory increases the average score from 3.971 to 4.105, with the most pronounced improvement in transition consistency, which rises from 3.252 to 3.671.
Removing the critic decreases the average score from 4.149 to 3.957. These results support the value of feedback-guided refinement
Disabling the experience library reduces the average score from 4.149 to 3.993, which suggests that experience provides useful guidance.
If VOL is replaced with a natural-language planner and the system follows a predefined workflow without loading any skills, the performance also decreases.

% \begin{figure*}
%     \centering
%     \includegraphics[width=\linewidth]{figures/shots_num_vs_results.pdf}
%     \vspace{-6mm}
%     \caption{\small The performance in different metrics with increasing shot numbers. VOL-Agent maintains stable performance. }
%     \label{fig:shot_nums}
%     \vspace{-2mm}
% \end{figure*}

% \begin{table*}
%     \setlength{\tabcolsep}{12pt}
%     \centering
%     \resizebox{\textwidth}{!}{
%     \setlength{\tabcolsep}{9pt}
%     \begin{tabular}{@{}ccccccccc@{}} 
%     \toprule
%     \multirow{2}{*}{\textbf{Methods} } & \textbf{Entity} & \textbf{Scene} & \textbf{Transition} & \textbf{Semantic} & \textbf{Cinematic} & \textbf{Narrative} & \textbf{Action\&} & \multirow{2}{*}{\textbf{Average}}\\
%     ~ & \textbf{Consistency} & \textbf{Consistency} & \textbf{Consistency} & \textbf{Alignment} & \textbf{Aesthetics} & \textbf{Coherence} & \textbf{Staging} \\
%     \midrule
%     w/o Critic & \\
%     w/o Experiment Book & \\
%     \textbf{VOL-Agent} &  \\
%     \bottomrule
%     \end{tabular}}
%     \label{tab:main_exp}
    
% \end{table*}

\subsection{Analysis}
\noindent\textbf{The effect of refinement.}
In VOL-Agent, we employ an iterative refinement strategy, as stated in Eq.\ref{eq:refinement}. Tab.~\ref{tab:BoN} presents the performance of iterative refinement with best-of-$N$ selection on a subset of our dataset.
Notably, even with multiple calls to the video model, best-of-$N$ still underperforms the refinement strategy, and scaling up to $N=6$ also fails to bridge this gap.

\noindent\textbf{The effect of director skill crafting.} To evaluate the director skills via self-crafting, we ask Gemini-3.8-flash to rank the results from different rounds and results obtained by directly taking the reference video as input or no reference video, according to their alignment with the reference video. Tab.~\ref{tab:director} shows that the director skills achieve a better alignment with the reference video's directing style.

\begin{figure}
    \vspace{-5mm}
    \centering
    \includegraphics[width=1.0\linewidth]{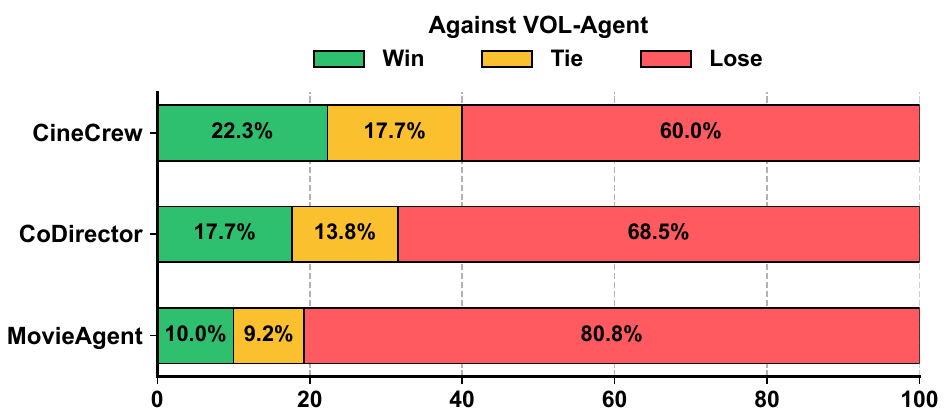}
    \vspace{-5mm}
    \caption{\small Human studies on different video agents against the proposed VOL-Agent.}
    \vspace{-5mm}
    \label{fig:human_studies}
\end{figure}

\noindent\textbf{Human Studies.}
To comprehensively evaluate the quality of generated videos, we conduct a rigorous human study following established protocols. Five professional evaluators with backgrounds in film production holistically rank the methods (VOL-Agent, CineCrew, MovieAgent, Co-Director) from best to worst, with ties allowed.
As shown in Fig.~\ref{fig:human_studies}, VOL-Agent outperforms the others.

\vspace{-2mm}
\section{Conclusion}
\vspace{-1mm}
In this paper, we presented VOL-Agent, a programmable framework for long-form video generation centered on the Video Orchestration Language. By separating global declaration and static validation from local runtime execution, VOL-Agent makes film structure explicit while preserving flexibility in tool and skill selection.
The structural memory mechanism introduced in VOL-Agent preserves character and scene consistency while ensuring the world state is carried over from the previous shot.
VOL-Agent is also capable of crafting director skills to make the workflow more similar to specified films.
Critic-guided refinement and experience reuse further support the correction and retention of production knowledge. Experiments on three benchmarks show strong overall performance, with additional gains from clay rendering via LLM-driven Blender. 
% Nevertheless, final outputs remain dependent on the capabilities of the underlying generators and may deviate from planned or rendered references. 
% Future work will investigate tighter correspondence between intermediate scene representations and generated videos, and evaluate continuity over longer and more complex narratives.

% \input{sec/6_llm_usage}

{
    \small
    \bibliographystyle{ieeenat_fullname}
    \bibliography{main}
}

\clearpage
\setcounter{page}{1}
\maketitlesupplementary

\section{Extended Related Work}
\vspace{-2mm}
\label{sec:related_full}

\noindent\textbf{Video Generation.}
% Since text-to-video models~\cite{makeavideo,emuvideo,svd,animatediff,videopoet,cogvideox,lumiere} have advanced rapidly.
% More recently, large-scale video foundation models like Sora~\cite{sora}, Veo-3~\cite{}, Kling~\cite{}, HunyuanVideo~\cite{}, Wan~\cite{}
Text-to-video (T2V) generation has advanced rapidly from early methods that apply image priors ~\citep{singer2022make, blattmann2023stable, hong2022cogvideo} to large-scale video foundation models such as CogVideoX~\citep{yang2025cogvideox}, HunyuanVideo~\citep{kong2024hunyuanvideo, wu2025hunyuanvideo}, Wan~\citep{wan2025wan} and MiniMax-H3~\citep{minimax-h3}, which have pushed generation quality, duration, and physical plausibility toward production level.
Furthermore, in controllable video generation, some approaches move beyond adapter-based extensions to existing models~\citep{jiang2025vace, guo2024sparsectrl, ICLR2024_94894cf9} and instead pursue omni-level video generation models. For example, proprietary models like Sora-2~\citep{Sora2}, Seedance-2.0~\citep{seedance2026seedance}, Kling-3.0-omni~\citep{team2025kling}, and Veo-3.1~\citep{veo3}, support conditioning on reference images or videos. Other methods, such as UniVideo~\citep{wei2025univideo}, VINO~\citep{chen2026vino}, Omni-Weaving~\citep{pan2026omniweaving}, and CogOmniControl~\citep{yang2026cogomnicontrol}, leverage the reasoning capabilities of VLMs to better interpret inputs.
Nevertheless, these models are fundamentally short-video generators, which typically produce clips on the scale of several seconds and no long-term video across the whole script. This gap motivates agent-based video generation, which aims to orchestrate multiple generation and coordinate them with script-level planning and long-term consistency management.
The integration of long-term memory mechanisms is also a crucial frontier for video world models~\cite{li2026forgewmprogressivecausaltraining}, enabling them to maintain a world state.

\noindent\textbf{Agentic Video Generation.}
The emergence of video agents~\cite{shi2025animaker, eskandar2026videoweaver} augments video generation with agentic planning, iterative refinement~\cite{liu2026cineforge}, and memory mechanisms.
For example, MovieAgent~\citep{wu2025automated} applies hierarchical planning, ViMax~\citep{huang2026vimax} builds a multi-agent system. CineCrew~\citep{chen2026better} establishes a FilmDSL, demonstrating the feasibility of coding video agents.
Unlike FilmDSL, our VOL-Agent focuses on declaring global cinematic structures and dependencies without prescribing how each shot should be generated, deferring the dynamic planning of concrete workflows, tools, and skills to runtime execution.
Test-time scaling~\citep{snell2025scaling} improves generation quality by investing additional computation during inference. For video test-time scaling, current methods focus on best-of-N sampling and prompt optimization~\citep{gao2025devil, hong2026comic, huang2025planning, long2026vista, zhu2026brandfusion}.
A key characteristic inherited from LLM agents is the memory mechanism~\citep{hu2025memory,zhang2025survey,zhou2026medical,wang2026opisinputgroundedbenchmarkmultiobject,zhou2026videomemory}, which enables an agent to preserve information across multiple interactions and sequential decisions.
Besides, long-form video generation typically requires multiple calls to video generation models to produce individual video segments. Memory is required to maintain story-level and shot-level continuity across these segments.

\section{Setup}

\subsection{Settings}
\label{sec:full_settings}
\noindent\textbf{Benchmarks.}
We conduct experiments on multi-scene long video generation. We collect these scripts from an existing open-source scripts website, which contains 20 scripts requiring recurring entities, multiple locations, and sequences of related actions.
We also conduct experiments on two public benchmarks, MovieBench~\cite{wu2025automated} and ViMax-Bench~\cite{huang2026vimax}. Note that we follow CineCrew to randomly sample 20 narratives from MovieBench.

\noindent\textbf{Setup.}
The comparison configuration specifies GPT-6.1-Sol~\cite{gpt-6.1-sol} for LLM planning and orchestration, and Gemini-3.7-flash~\cite{gemini-3.7-flash} for VLM, GPT-Image-2~\cite{gpt-image-2} for image generation, and MiniMax-H3~\cite{minimax-h3} for video generation.
For a fair comparison, all compared methods in this paper employ the same LLM/VLM, image models, and video models as VOL-Agent. Meanwhile, we adopt the same critic agent as VOL-Agent across all compared methods to perform Best-of-$N$ selection (n=2) under the same computational budget.
The resolution of video generation is set to 720P.

\subsection{Metrics}
\label{sec:full_metrics}
We evaluate seven dimensions scored independently on an integer scale from 0 to 5: Entity Consistency, Scene Consistency, Transition Consistency, Semantic Alignment, Cinematic Aesthetics, Narrative Coherence, and Action \& Staging.
Entity consistency compares visible identities with the supplied state-appropriate references. Scene consistency checks the location and its landmarks. Semantic alignment measures whether the script-derived beat occurs. Cinematic aesthetics covers composition, camera work, lighting, and materials; narrative coherence evaluates the causal relationships required by the beat; and action and staging evaluates visible movement and interactions. These dimensions are reported separately to avoid hiding a continuity failure behind an aesthetic score.
The independent VLM evaluation is configured to use Gemini-3.8-Flash~\cite{gemini-3.8-flash}.

\begin{figure*}[!ht]
    \vspace{-4mm}
    \centering
    \includegraphics[width=\linewidth]{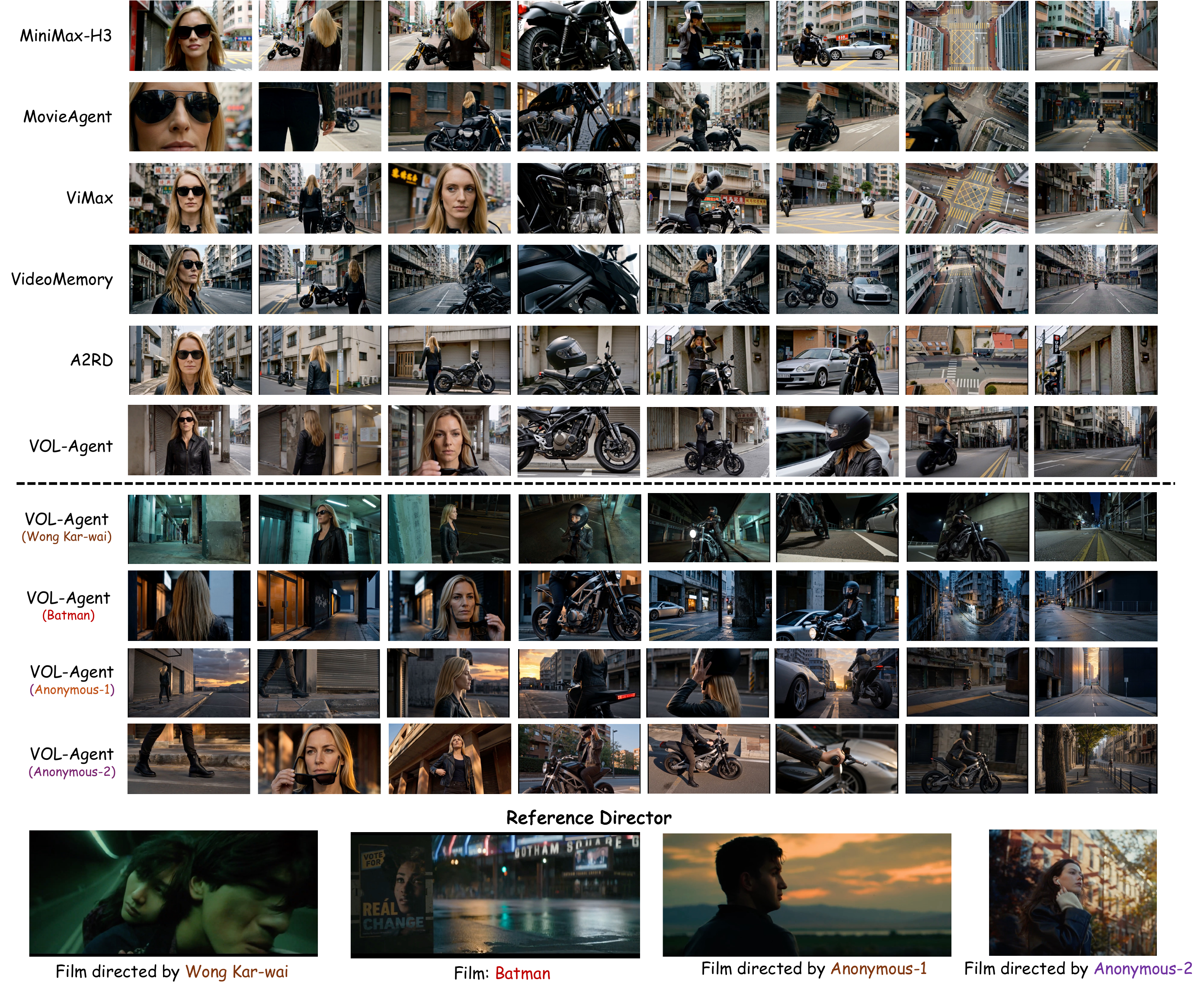}
    \vspace{-6mm}
    \caption{\small Additional visual comparisons and director-style examples. The upper rows compare video agents; the lower rows show VOL-Agent outputs conditioned on filmmaking styles, with reference films shown below.}
    \label{fig:visual_result}
    \vspace{-4mm}
\end{figure*}

\subsection{Baseline Implementation}

\noindent\textbf{Autoregressive Base Models.}
For a fair comparison, the methods compared are implemented in the form of an autoregressive paradigm. We use the same LLM and VLM as VOL-Agent to plan each shot, and the same image model to generate entity and scene reference images. Each shot generation uses the previous shot as a reference video input.

\noindent\textbf{Video Agentic System.}
The compared methods, such as MovieAgent, ViMax, VideoMemory, CineCrew, Co-Director, and ScriptAgent, applied the same LLM (GPT-6.1-Sol), VLM (Gemini-3.7-flash), image model (GPT-image-2) and video model (MiniMax-H3). We required them to perform Best-of-$N$ with N set to 2, while VOL-Agent required one round of refinement, meaning both were required to call the video model twice.

\section{Additional Qualitative Results}
\label{sec:additional_visual}

In Fig.~\ref{fig:visual_result}, we present a comparison of VOL-Agent with different methods. Due to the lack of long-term memory, the autoregressive MiniMax-H3 shows that scene transitions and layout vary significantly across different generation rounds.
While MovieAgent demonstrates strong cinematography, it similarly suffers from a lack of memory and material reference capabilities. For instance, the rider's helmet is missing during the ride, along with incorrect scene and character configurations.
Moreover, video agents with long-term memory mechanisms, such as ViMax, VideoMemory and A${}^2$RD, fall into confusion regarding scene adherence. The same scene repeatedly appears across different shots, creating a narrative illusion that the protagonist is constantly riding in the same location.
The lower rows also illustrate director-style variations using the reference films shown in Fig.~\ref{fig:visual_result}.

\section{Benchmark}
\vspace{-1mm}
The experiment of VOL-Agent is conducted on our benchmark and two public benchmarks. Our benchmark is collected from existing film scripts and will be made public, which contains 20 long stories.
In this benchmark, VOL-Agent generated videos with a total duration of approximately 4,910 seconds, with each video containing about 22 shots.

\begin{figure*}
    \vspace{-3mm}
    \centering
    \includegraphics[width=1.0\linewidth]{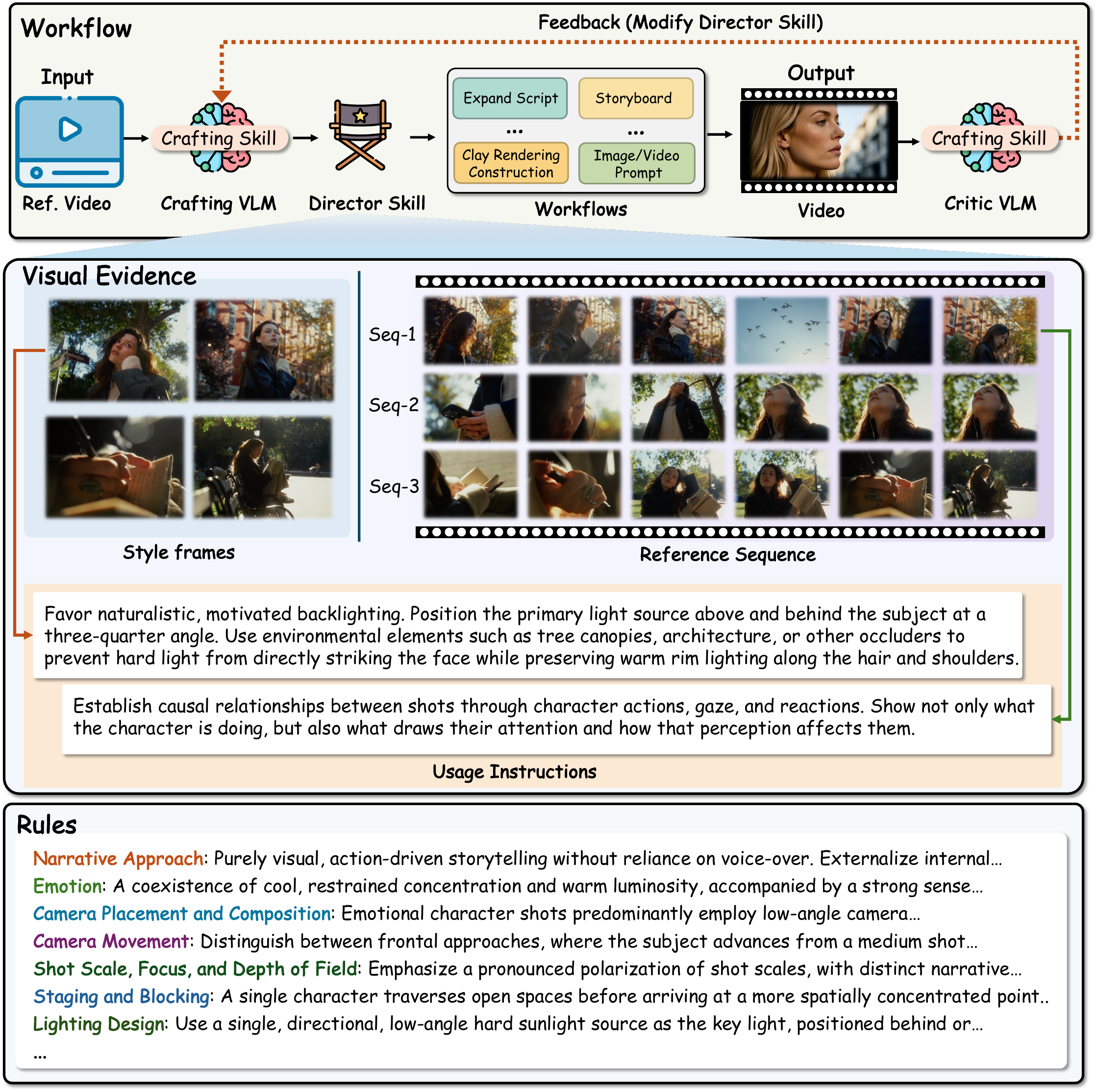}
    \vspace{-7mm}
    \caption{\small The Construction of Director Skill. }
    \label{fig:director}
    \vspace{-4mm}
\end{figure*}

\section{Director Skill via Crafting}
\vspace{-2mm}
Fig.~\ref{fig:director} illustrates how VOL-Agent constructs and applies a director skill from a reference video. 
VOL-Agent uses a Crafting Skill to a tool in analyzing the reference, organizing visual evidence, and producing a reusable director skill. The resulting skill provides both filmmaking rules and evidence-grounded instructions for their application. 
The visual evidence contains Style frames and Reference sequences. 
\textbf{Style frames} are individual reference frames selected to illustrate a photographic relationship, such as composition, lighting, colour, or material appearance. \textbf{Reference sequences} are ordered frame lists that support temporal and narrative organization. For example, a sequence can direct the author to consider how a character’s action, gaze, and reaction guide the viewer’s attention across successive observations. 
Other learned rules organize these instructions into a director skill covering camera placement, movement, shot scale, staging, and lighting, as illustrated in the lower part of Fig.~\ref{fig:director}
It is then made available to the production workflows. After video generation, a critic VLM examines the output video and then provides feedback for crafting VLM, thus improving the director skill.

\onecolumn  
\section{Domain Specified Language}
The VOL represents a film through reusable entities, scenes, and video segments. Each shot declares one segment, including its beat, duration, visible entities, and input dependencies. Example~\ref{lst:vol-example} shows the structure of \program. 
% The segment lists declare visible entities.
% The first segment defaults to \texttt{cut}, and the following segment

% \begin{minipage}

\vspace{-4mm}
\begin{lstlisting}[language=Python,frame=single,basicstyle=\ttfamily\footnotesize,breaklines=true,columns=fullflexible,keepspaces=true,showstringspaces=false,float=tp,captionpos=b,caption={A minimal VOL declaration.},label={lst:vol-example}]
film.title = "The Knight"
film.global_script = (
    "A knight in armor riding a horse enters a castle, and after being knighted, heads to a muddy swamp."
)
film.video_model = "MiniMax-H3"
film.follow_video_model = "MiniMax-H3"

knight = film.entity(
    "knight", "A European knight around 40 years old, with a beard, blond hair, and a resolute gaze.",
    voice="Deep, restrained baritone voice, steady pacing",
    states={
        "armored": "Intact dark gray iron armor, longsword hanging at the waist, worn scratches on the breastplate, riding boots stained with road dust",
        "battle_worn": "Iron armor stained with soot and scorch marks, damaged right greave, ashes on the cheek, sword-holding arm trembling slightly",
    }
)
horse = film.entity(
    "horse", "A war horse in armor."
)

with film.scene(
    "city_bridge", setting="Medieval royal city drawbridge entrance, shrouded in morning mist. Acting area: The main drawbridge roadway is about four meters wide, paved with bluestone with moss growing between the cracks, echoing crisply beneath horse hooves.",
    entities=[knight, horse]
) as sc:
    s1 = sc.shot(
        "beat-1: knight rides his horse."
        "beat-2: The knight rides up to the iron chain bollards at the bridgehead, slows his horse's pace, and looks up toward the city gate archway.",
        duration=12, edge=cut, entities=[knight, horse]
    )
    s2 = sc.shot(
        "beat-1: Close-up of the knight's face.",
        duration=4, edge=cut, entities=[knight])
    )

with film.scene(
    "palace_hall", setting="Main hall of the palace, high vaulted ceiling, stone pillars lined up along both sides, their surfaces showing age-old wear and cracks.",
    entities=[knight, king]
) as sc:
    s1 = sc.shot(
        "The king sits on the throne, holding a scepter in his hand."
        duration=4, edge=cut, entities=[king]
    )
    s2 = sc.shot(
        "The king stands up, walks down the steps, and slowly approaches the knight."
        duration=6, edge=follow, entities=[king, knight]
    )
    
\end{lstlisting}

\twocolumn
\clearpage

\end{document}